%% file: acl_latex.tex
\pdfoutput=1

\documentclass[11pt]{article}

\usepackage[preprint]{acl}

\usepackage{times}
\usepackage{latexsym}
\usepackage{enumitem}
\usepackage[T1]{fontenc}

\usepackage[utf8]{inputenc}

\usepackage{microtype}

\usepackage{inconsolata}
\usepackage{url}
\usepackage{microtype}
\usepackage{booktabs}
\usepackage{multirow} 
\usepackage{subcaption}
\usepackage{amsmath}
\usepackage{amssymb}
\usepackage{amsfonts}
\usepackage{mathtools}
\usepackage{mathrsfs}
\usepackage{amsthm}
\usepackage{letltxmacro}
\usepackage{algorithm}
\usepackage{algorithmic}
\usepackage[english]{babel}
\usepackage{amsthm}
\usepackage{array}
\usepackage{listings}

\title{SA-MDKIF: A Scalable and Adaptable Medical Domain Knowledge Injection Framework for Large Language Models}

\author{Tianhan Xu$^{1,2}$ \,  Zhe Hu$^{3}$ \, Ling Chen$^{1,2}$ \, Bin Li$^{1,2}\thanks{~~Corresponding author.}$
\\
$^{1}$School of Information Engineering, Yangzhou University\\
$^{2}$Jiangsu Province Engineering Research Center of Knowledge Management and Intelligent Service\\
  $^{3}$Department of Computing, The Hong Kong Polytechnic University
 \\
  $^{1,2}${\tt  \{xutianhan2018@gmail.com,lchen@yzu.edu.cn,lb\_kmis@yzu.edu.cn\}},
  $^{3}${\tt zhehu.94@gmail.com} 
  }

\begin{document}
\maketitle
\begin{abstract}
Recent advances in large language models (LLMs) have demonstrated exceptional performance in various natural language processing (NLP) tasks. However, their effective application in the medical domain is hampered by a lack of medical domain knowledge. In this study, we present SA-MDKIF, a scalable and adaptable framework that aims to inject medical knowledge into general-purpose LLMs through instruction tuning, thereby enabling adaptability for various downstream tasks. SA-MDKIF consists of two stages: \textbf{skill training} and \textbf{skill adaptation}. In the first stage, we define 12 basic medical skills and use AdaLoRA to train these skills based on uniformly formatted instructional datasets that we have constructed. In the next stage, we train the skill router using task-specific downstream data and use this router to integrate the acquired skills with LLMs during inference. Experimental results on 9 different medical tasks show that SA-MDKIF improves performance by 10-20\% compared to the original LLMs. Notably, this improvement is particularly pronounced for unseen medical tasks, showing an improvement of up to 30\%.
\end{abstract}

\section{Introduction}
\input{intro}

\section{Related Work}
\input{related}

\section{Method}
\input{method}

\section{Experiments}
\input{experiment}

\section{Conclusion}
In this work, we propose a \textbf{S}calable and \textbf{A}daptable \textbf{M}edical \textbf{D}omain \textbf{K}nowledge \textbf{I}njection \textbf{F}ramework for LLMs (SA-MDKIF). For this purpose, we first design 12 medical skills and then train them with AdaLoRA in Stage I. Then, we train the skill router based on the specific downstream task data and use the router to fuse the medical skills with the general-purpose LLM in Stage II. Our approach allows for the integration of medical domain knowledge while maintaining the general capabilities of the LLM.

Extensive experiments on 9 downstream datasets show that SA-MDKIF significantly improves the performance of the original LLM and achieves state-of-the-art performance on the downstream medical tasks. Moreover, SA-MDKIF can handle the few-shot learning settings and produce surprising results on unseen tasks. It is worth noting that our framework can be applied to other domains besides medicine.

\input{reference}

\appendix

\label{sec:appendix}
\input{appendix}

\end{document}

%% file: intro.tex
Recent Large Language Models (LLMs), such as ChatGPT~\cite{chatgpt}, Llama 2~\cite{llama2}, and PaLM 2~\cite{palm2}, have gained significant attention due to their impressive performance, robust generalization, and reasoning capabilities across various Natural Language Processing (NLP) tasks, such as question answering, text summarization, and natural language inference~\cite{Qin2023IsCA}. 
However, when it comes to applying these models directly to medical NLP tasks, there are inherent challenges.
Despite their strong general capabilities, the general pre-training of LLMs lacks the specialized domain knowledge necessary for effectively tackling practical clinical challenges. Medical Tasks such as ICD coding~\cite{mullenbach2018explainable}, medication recommendation~\cite{jensen2012mining}, and readmission prediction~\cite{shulan2013predicting}, require an in-depth understanding of medical intricacies that goes beyond what general pre-training offers. This highlights a gap in their application in the medical field.

\begin{figure}[t]
    \centering
    \includegraphics[width=0.5\textwidth]{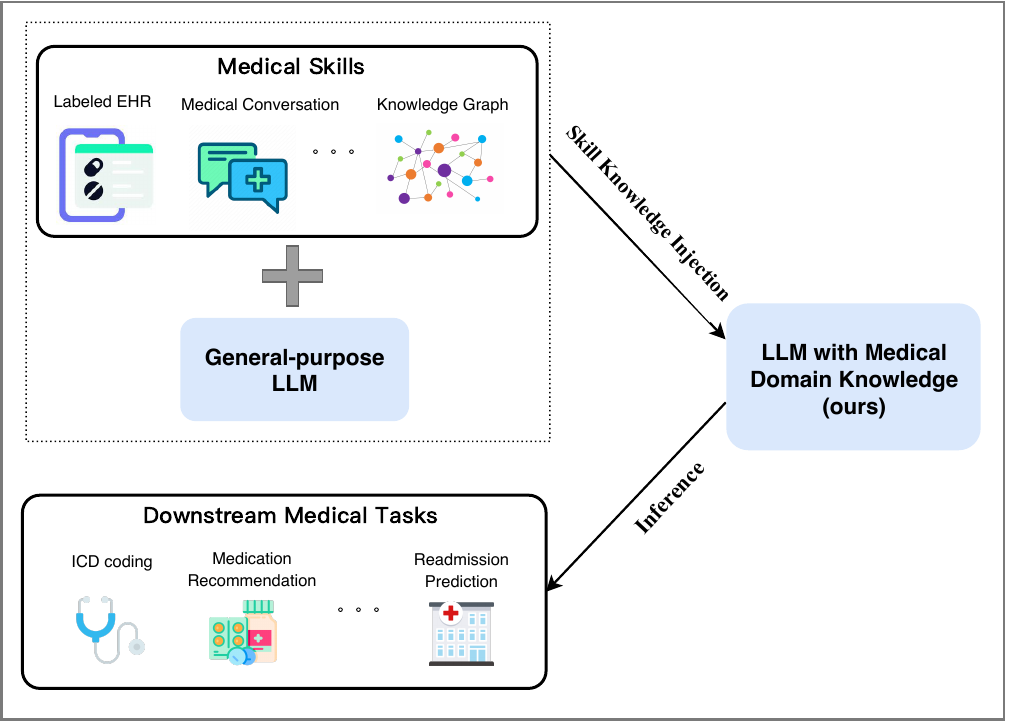}
    \caption{\small The goal of this work is to inject medical knowledge into the general-purpose LLM for its adaptation to different downstream medical tasks.}
    \label{fig1}
    \vspace{-4mm}
\end{figure}

To address this challenge, previous efforts have approached the problem by continually training the general-purpose LLMs with medical data to incorporate domain knowledge. 
One line of research focuses on continuous pre-training the LLMs with unsupervised medical corpora~\cite{Med-PaLM, Peng2023ASO, chen2023meditron}. However, such unsupervised pretraining is resource-intensive and time-consuming~\cite{ding2023parameter}, and may lead to catastrophic forgetting that can weaken the  generalization capabilities of the model~\cite{luo2023empirical}. 
Alternatively, other efforts concentrate on fine-tuning LLMs using medical data through instruction tuning techniques~\cite{wang2023huatuo, LLM4EHR, li2023chatdoctor}. Nevertheless, they lack flexibility in using knowledge to perform downstream tasks because they cannot determine which knowledge is essential for a particular downstream task. Task-specific fine-tuning has been shown to be more effective than multitask learning and generalized fine-tuning~\cite{raheja2023coedit}.

In this work, we propose SA-MDKIF, a novel \textbf{S}calable and \textbf{A}daptable \textbf{M}edical \textbf{D}omain \textbf{K}nowledge \textbf{I}njection \textbf{F}ramework for LLMs. SA-MDKIF can effectively adapt a general-purpose LLM such as Llama 2 \cite{llama2} to the medical domain, as shown in Figure 1. Our framework consists of two core stages: \textbf{upstream skill training} and \textbf{downstream skill adaptation}. In the first stage for skill training, we design 12 basic medical skills which are essential to solve a wide range of medical tasks. Then we adopt a skill-relevant instruction training to incorporate each skill into the LLM with a specific AdaLoRA modular~\cite{AdaLoRA}. 
Such skill-relevance instruction training improves training efficiency and enables a more flexible integration of knowledge into the LLMs, without changing the original LLM parameters.
In the second stage for downstream task fine-tuning, our famework uses a skill router-based method, which involves learning the weights of each skill through gradient descent or CMA-ES~\cite{hansen1996adapting}, enabling the dynamic combination of these skills to effectively tackle specific medical tasks during inference. This two-stage process ensures that our model is not only efficient in its skill learning process but also versatile in applying its learned skills to collaboratively solve various medical tasks.

To evaluate our model performance, we conduct extensive experiments on
9 downstream medical tasks including 3 unseen tasks: ICD coding, medicine recommendation, and readmission prediction. Experimental results show that our method significantly outperforms the original Llama 2 in 9 downstream medical tasks, with corresponding metrics improved by 10-30\%. Furthermore, our method achieves the state-of-the-art of these medical tasks in both normal and few-shot settings, demonstrating its robustness and adaptability in various real scenarios.

To sum up, our work has the following contributions:

\begin{itemize}[nolistsep]
\item We present SA-MDKIF, a two-stage scalable and adaptive medical domain knowledge injection framework that can tackle different downstream medical tasks in both normal and few-shot settings.
\item We design 12 basic medical skills and use AdaLoRA to train their parameters. Then we adaptively fuse the skills with the general-purpose LLM by skill router based on the specific downstream task. 
\item Experimental results show that SA-MDKIF can dramatically outperform the original LLMs in medical tasks, especially in unseen tasks. Moreover, the performance of our method surpasses the baseline models in the downstream medical tasks.
\end{itemize}

%% file: related.tex
\subsection{BERT-based Medical Models}
Some previous work~\cite{Li2022ChineseEM,hu-etal-2022-mocha,aribandi2022ext} pre-trained token representation on a medical corpus based on BERT~\cite{bert}, and then fine-tuned downstream task data based on the representation of input tokens. BioBERT~\cite{lee2020biobert} is continuously trained using PubMed abstracts and PMC full-text articles on top of the generalized corpus pre-training, thus injecting biomedical knowledge at the pre-training stage. SMedBERT~\cite{zhang-etal-2021-smedbert} simultaneously introduces the medical entities in the knowledge graph, together with the structured semantic information in the entity relationships, into the pre-trained model. G-BERT~\cite{ijcai2019p825} combines GNNs and BERT to learn medical code representations of hierarchies and further integrates the results into pre-trained Transformer-based models. MedM-PLM~\cite{MedM-PLM} explores the interaction of structured and unstructured data by learning enhanced electronic health records(EHR) representations through pre-training tasks that correlate these two modalities. Yang et al. proposed GatorTron~\cite{yang2022large}, a PLM for EHR, and achieved accurate results on five medical NLP tasks.

However, such models have limitations. First, these models are trained on a single type of medical corpus, which affects their contextual understanding and reasoning; second, these models are limited in model scale, and since they are designed based on BERT, their ability of few-shot learning is insufficient, and they have poor performance when confronted with unseen medical tasks.

\subsection{LLMs in the Medical Domain}
Recently, generative large language models have shown strong generalization and few-shot learning capabilities in various tasks~\cite{brown2020language}. Therefore, some researchers have considered training LLMs on medical corpus. Google and Deepmind introduce prompt tuning based on their multi-category medical datasets, and train the medical LLMs that called Med-PaLM 2~\cite{med-palm2} from scratch. GatorTronGPT~\cite{Peng2023ASO}, a clinical large language generation model, can be used for biomedical natural language processing, clinical text generation and evaluation. It uses a unified P-tuning~\cite{Liu2022PTuningPT} base text generation architecture to address biomedical relationship extraction and question answering. PMC-LLaMA~\cite{wu2023pmc} and Huatuo~\cite{wang2023huatuo} are based on LLaMA~\cite{touvron2023llama} as the original LLM and then fine-tuned using medical papers and knowledge graphs, respectively. DoctorGLM~\cite{xiong2023doctorglm} is based on ChatGLM~\cite{zeng2022glm} and fine-tuned with Chinese medical dialog data. 

However, training the aforementioned medical LLMs requires a large corpus and consumes significant time and memory resources. In addition, updating and extending the corpus of these LLMs is a challenging task.

\subsection{PEFT}
Parameter-efficient fine-tuning (PEFT)~\cite{ding2023parameter} fine-tunes only a small subset of additional model parameters, leaving most LLM parameters fixed. PEFT significantly reduces computational and storage costs and can achieve accuracy comparable to full parameter fine-tuning.

\noindent\textbf{Addition-based:} Adapter-tuning~\cite{He2021OnTE} introduces small-scale neural network modules (Adapter) between Transformer sublayers as fine-tuning parameters. Prompt-tuning~\cite{lester-etal-2021-power} and P-tuning~\cite{Liu2022PTuningPT} perform model fine-tuning with trainable, parameterized prompts.

\noindent\textbf{Specification-based:} BitFit~\cite{zaken2021bitfit} achieves parameter reduction by training only the bias-terms and task-specific classification layer in the original model while freezing other parameters.

\noindent\textbf{Reparameterization-based:} LoRA~\cite{hu2021lora} reduces the number of training parameters through a low-rank matrix representation, enabling efficient fine-tuning of LLMs with a small number of parameters. Its improved variant, QLoRA~\cite{dettmers2023qlora} achieves approximate computation through a frozen 4-bit quantized PLM. 

Our work is based on the reparameterization-based methods. The reason is that, compared with addition-based methods, reparameterization methods do not need to insert additional neural network modules, and have better inference performance and convergence speed; moreover, compared with specification-based methods, reparameterization methods have better performance~\cite{delta-tuning}.

\subsection{Mixture of Experts}
The Mixture of Experts (MoE) framework, originally proposed by Masoudnia and Ebrahimpour~\cite{MoE-1}, has become central to capturing complex relationships in diverse data sets. It employs specialized expert subnetworks that are dynamically controlled by a gating network, allowing for efficient adaptation to varying data structures. In the area of language model scaling, Du et al. introduced GLaM~\cite{MoE-2}, demonstrating the efficiency of MoE in improving large language models. In addition, Wang et al. introduced Adamix~\cite{wang2022adamix}, a mixture-of-adapters approach that highlights the versatility of MoE in optimizing language models. To deal with data conflicts during instruction fine-tuning, Chen et al. proposed LLaVA-MoLE~\cite{chen2024llava}, which uses a sparse mixture of LoRA experts. Taken together, these studies highlight the adaptability and effectiveness of MoE in various machine learning scenarios. Our framework with skill-relevant parameters can be regarded as an MoE system which flexibly combines skills to jointly solve the downstream task.

%% file: method.tex
\begin{figure*}
    \centering
    \includegraphics[width=1.0\textwidth]{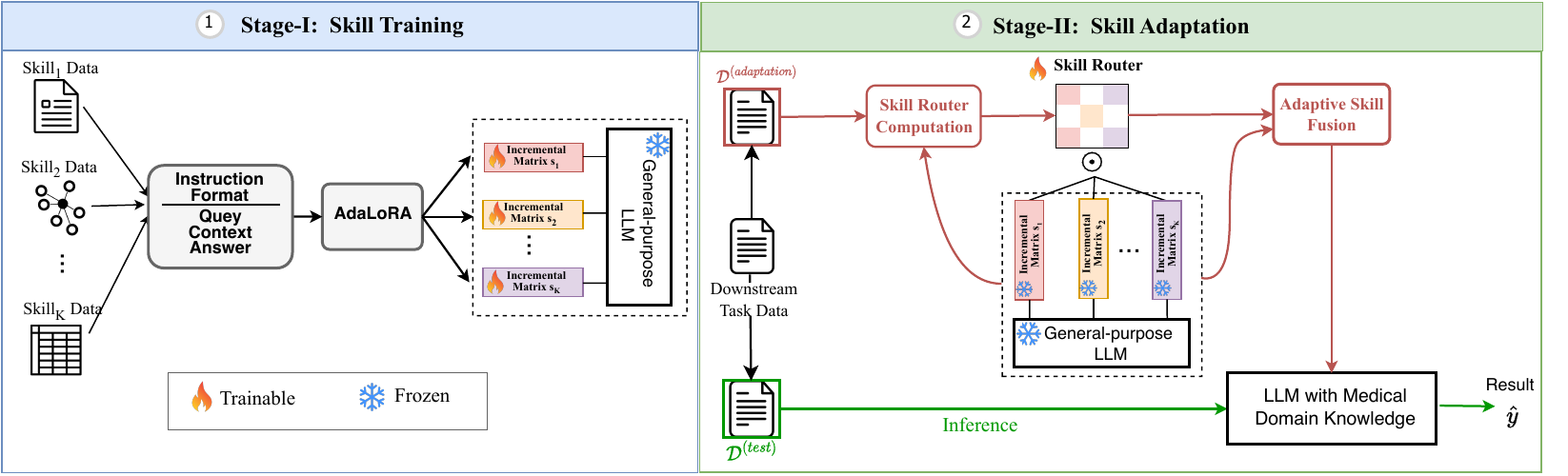}
    \caption{\small The framework of SA-MDKIF. It consists of two stages: skill training and skill adaptation. In the skill router computation of Stage-II, we use gradient descent and CMA-ES to compute the router for normal settings and few-shot settings, respectively. The red and green lines represent the adaptation and inference processes, respectively.}
    \label{fig2}
\end{figure*}

\subsection{Overview}
Our proposed framework is illustrated in Figure 2. SA-MDKIF consists of two stages: (1) Construction and training of medical skills. (2) Skill adaptation based on the downstream task and fuse adaptive skills with the general-purpose LLM. In this way, our model acquires the generalized abilities of LLM as well as the specific knowledge of medical skills. In the following, we describe our method in detail.

\subsection{Skill Training Stage}
\subsubsection{Skill Design}
First, we design 12 types of knowledge in medical domain that cover the basic skills for solving medical NLP tasks. They include: Question Answering (QA) Skill, Multiple Choice Question Answering (MCQA) Skill, Medical Conversation(MC) Skill, Multi-Label Document Classification (MLDC) Skill, Machine Reading Comprehension (MRC) Skill, Natural Language Inference (NLI) Skill, Text Summarization (TS) Skill, Named Entity Recognition (NER) Skill, Relation Extraction (RE) Skill, Entity Attribute (EA) Skill, Entity Synonymy (ES) Skill , Entity-Entity Relation (ER) Skill. 

All of these skills are necessary for downstream medical tasks. For example, similar medical sentences can be learned through the NLI skill, appropriate medical labels can be assigned to EHR through the MLDC skill, and more disease and medical attributes and relations can be learned by the model through the EA and ER skills. It is worth mentioning that we also include causality in the RE skill, which is crucial for disease analysis, such as disease etiology, risk factors, comorbidities, and so on. 

In the following we use $\mathcal{B} =\{s_1,s_2,...,s_K\}$ to describe these skills, where $K$ is the number of skills.

\begin{figure}[htbp]
    \centering
    \includegraphics[width=0.48\textwidth]{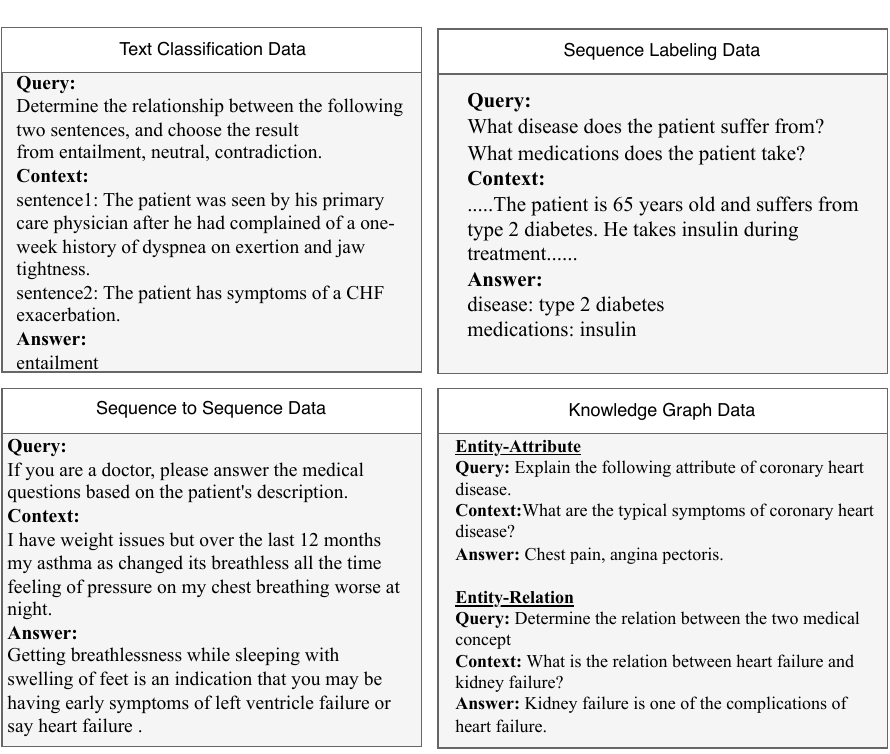}
    \caption{Examples of converting several categories of medical data to the unified instruction format.}
    \label{fig3}
\end{figure}

\begin{algorithm*}
\caption{Skill Training Stage}\label{algo1}
\begin{algorithmic}[1]
    \STATE{{\bfseries Input:} K medical skill datasets $\mathcal{D}_1^{(train)}, \dots, \mathcal{D}_K^{(train)}$; \par
    weight matrix of the original LLM $W^{(0)}$; \par
    initial fine-tuning warm-up step $t_0$, final fine-tuning step $t_1$. } \\
    \FOR{$1 \leq i \leq K$} 
        \STATE Incremental matrix parameterization by (\ref{equ:svd}).
        \FOR{$t_0 \leq t \leq t_1$} 
            \STATE Sample a mini-batch from $\mathcal{D}_i$ and update gradient by (\ref{equ:gradient_update}).
            \STATE Eigenvalue gradient trimming by (\ref{equ:gradient_trimming}).
        \ENDFOR
    \ENDFOR
    \STATE{{\bfseries Output:}The trained parameters of medical skills $\Theta = \{\theta_1, \theta_2, ..., \theta_K\}$ ,
    where $\theta_i = \{W_q^i, W_k^i, W_v^i, W_{f_1}^i, W_{f_2}^i, W_o^i\}$.}
\end{algorithmic}
\end{algorithm*}

\subsubsection{Unified Data Format}
In order to construct the above 12 skills, we selected 12 public medical datasets with different contents for training. The heterogeneous data of the 12 skills are classified into 5 categories: question answering data, text classification data, sequence labeling data, sequence to sequence data, knowledge graph data. We uniformly convert these categories into the \textbf{instruction format}. The format takes a \textit{\textbf{context}} and a \textit{\textbf{query}} as input, and the \textit{\textbf{answer}} to the query as output. Figure 3 shows the process of converting several categories of data into instruction format. \\
(1) \textbf{Text Classification Data.} For TC, MLDC and NLI data, we use the original input text as context and then construct a query containing all valid labels. The context and query are concatenated as input. The skill module is trained to extract the answers in the query by predicting the start and end positions of the query. \\
(2) \textbf{Sequence Labeling Data.} For NER, MRC and causal discovery data, we manually design task-specific templates to map inputs to the desired contexts and queries. \\
(3) \textbf{Sequence to Sequence Data.} For medical conversations and text summary data, contexts and answers are input and output sequences, respectively, and queries are generated based on task templates. \\
(4) \textbf{Knowledge Graph Data.} The medical knowledge graph contains two types of data, entity descriptions and entity relations. We add two different queries to them and train the model to output the description of an entity or to predict the relation between two entities.

\subsubsection{Training Methods}
We follow the efficient reparameterization method AdaLoRA~\cite{AdaLoRA} to train each skill $s_i$. In the skill training stage, the amount of LLM parameter changes is modeled by a low-rank decomposition, so that parameter tuning of LLM is achieved with very small incremental parameters. Compared to LoRA~\cite{hu2021lora}, AdaLoRA tunes more layers in the transformer block, while dynamically adjusting the rank of each incremental matrix according to its importance to achieve better performance. 

We parameterize the incremental matrix update $\Delta \in \mathcal{R}^{d_1 \times d_2}$ of the original LLM weight matrix $W^{(0)}$ in singular value decomposition form:
\begin{equation}\label{equ:svd}
   W=W^{(0)} + \Delta \approx W^{(0)} + U\Lambda V
\end{equation}
where $U \in \mathcal{R}^{d_1 \times r}$ and $V \in \mathcal{R}^{r \times d_2}$ donate the left and right sigular vectors of $\Delta$. $\Lambda \in \mathcal{R}^{r \times r}$ is the diagonal matrix with the singular values as its elements. Since $r \ll min(d_1, d_2)$, the number of training parameters is greatly reduced compared to full fine-tuning. 
We use $W_q$, $W_k$, $W_v$ to refer to the query, key, and value matrices in the self-attention block, and define the weight matrices of two linear layers in the FFN as $W_{f_1}$ , $W_{f_2}$. The oputput projection is $W_{o}$. Then, we perform an SVD parameterization on each weight matrix containing $W_q$, $W_k$, $W_v$, $W_{f_1}$ ,$W_{f_2}$,$W_{o}$ of each transformer layer. The above steps is used to train the incremental matrix parameters of the entire $K$ skills in parallel on each dataset $\mathcal{D}_i^{(train)}$. After training, we get the skill parameter matrix set $\Theta = \{\theta_1, \theta_2, ..., \theta_K\}$
where $\theta_i = \{W_q^i, W_k^i, W_v^i, W_{f_1}^i, W_{f_2}^i, W_{o}^i\} , i \in \{1,...,K\} $. Appendix A describes the details of AdaLoRA. See Algorithm 1 for the main flow of training algorithm.

\subsection{Skill Adaptation Stage}
\subsubsection{Adaptation for Downstream Task}
In the skill adaptation stage, the trained skills are adaptively combined based on different downstream tasks for differentiated knowledge injection. For the downstream tasks, we use the same approach to convert them to the instruction format. Each task $\mathcal{T}$ is represented by the context-target pairs: ${(x_n, y_n)}_{n=1,...,N}$, where $x_n$ is the input sequence of tokens including query and context, and $y_n$ is its corresponding answer sequence of tokens, $N$ donates the total number of samples. 

Formulating the upstream skills and downstream tasks into a unified format has two advantages. (1) It can bridge the gap between the two stages in our framework. (2) The queries generated from the designed task-specific templates are semantically rich prompts that can better stimulate the potential of LLM for higher performance. 

We divide the downstream task data $\mathcal{D}$ into two parts: $\mathcal{D}^{(adaptation)}$ and $\mathcal{D}^{(test)}$, where $\mathcal{D}^{(adaptation)}$ for skill adaptation and $\mathcal{D}^{(test)}$ for test. Inspired by the previous work~\cite{wang2022adamix, chen2024llava}, we implement skill adaptation by the \textbf{Skill Router}, and compute the router $\mathcal{R}$ by the following formulas:

\begin{equation}\label{equ:e_matrix}
   E = A \cdot \Theta + \boldsymbol{b}
\end{equation}
where $E = [\boldsymbol{e_1}, \boldsymbol{e_2}, ..., \boldsymbol{e_K}]$ and $K$ is the number of skills. $A = [\boldsymbol{\alpha_1}, \boldsymbol{\alpha_2}, ..., \boldsymbol{\alpha_K}]$ donates the weight matrix of skills, $\boldsymbol{b}$ is the bias vector. $A$,$\boldsymbol{b}$ are trained using $\mathcal{D}^{(adaptation)}$. \\

\begin{equation}\label{equ:gate_function}
   \mathcal{R}(\boldsymbol{e}_i) = \frac{\exp\left( \boldsymbol{e}_i / {\tau}\right)}{\sum_{j=0}^{K} \exp\left( \boldsymbol{e}_j / {\tau}\right)}
\end{equation}
where $\tau$ is the temperature coefficient to smooth the output distribution. Given the parameters of all the above trained skills $ \Theta = \{\theta_1, \theta_2, ..., \theta_K\}$ and the frozen parameter $\Phi_0$ of original LLM, the objective function is to compute the matrix $A$ and vector $\boldsymbol{b}$ by maximizing the probability of generating the target sequence $y$. The loss function $\mathcal{L}_{task}$ of $\mathcal{T}$ is as follows:
\begin{equation}\label{equ:incrementalParams}
    \Delta \Phi = \sum_{i=1}^K \mathcal{R}(\boldsymbol{e}_i) \cdot \theta_i
\end{equation}
\begin{equation}\label{equ:adaptationObjective}
\mathcal{L}_{task} = - \sum_{(x,y) \in \mathcal{Z}} \sum_{t=1}^{|y|} log \left( p_{\Phi_0 + \Delta \Phi} (y_t | x, y_{<t}) \right )
\end{equation}
Matrix $A$ and vector $\boldsymbol{b}$ are randomly initialized, since $\Theta$ is trained in the first stage, we only need to iteratively update $A$,$\boldsymbol{b}$ during the adaptation stage. Our approach is parameter efficient because $ \Delta \Phi \ll \Phi_0$. 

To ensure stability of training, accuracy of results, and avoid overfitting, we categorize each downstream task $\mathcal{T}$ into one of the two settings based on its sample size: the \textbf{normal settings} and the \textbf{few-shot settings}.  

\begin{figure}
    \centering
    \includegraphics[width=0.5\textwidth]{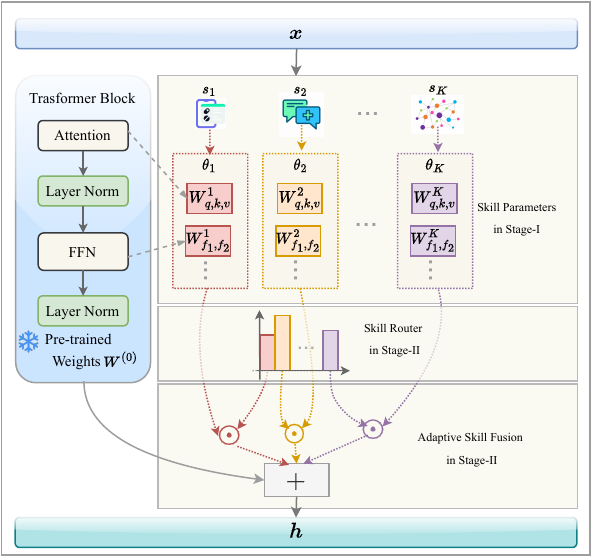}
    \caption{\small Details of our proposed fusion process.}
    \label{fig4}
\end{figure}

For the \textbf{normal settings}, we use the \textit{\textbf{gradient descent}} to minimize the loss function. To prevent the model from relying too much on a single skill, which leads to overfitting and thus affects the generalization performance, we introduce a regularity term to penalize excessively high coefficients $\boldsymbol{\alpha_i}$. 
\begin{equation}\label{equ:regularObjective}
   \mathcal{L}_r = \sum_{i=0}^K ||\boldsymbol{\alpha}_i||_2^2
\end{equation}
where $||\boldsymbol{\alpha}_i||_2^2$ donates the $L_2$ paradigm of $\boldsymbol{\alpha}_i$. The final loss function $\mathcal{L}$ can be described as follows:
\begin{equation}\label{equ:normalObjective}
   \mathcal{L} = \mathcal{L}_{task} + \gamma_1 \mathcal{L}_r
\end{equation}
where $\gamma_1$ serves as a hyperparameter. 

For the \textbf{few-shot settings}, we use the \textit{Covariance Matrix Adaptive Evolution Strategies \textbf{(CMA-ES)}} for skill adaptation. CMA-ES~\cite{hansen1996adapting} is an Evolutionary Algorithm designed to solve nonlinear, nonconvex, population-based optimization problems. CMA-ES is typically applied to problems with a search space dimension between 3 and 100. We use the CMA-ES algorithm in the few-shot settings to shape the search space, where the weight matrix $A$ and bias vector $\boldsymbol{b}$ are chosen on the basis of their performance on the few-shot samples of $\mathcal{D}^{(adaptation)}$. 

\begin{algorithm*}
	\caption{Skill Adaptation Stage} 
	\begin{algorithmic}[1]       
        \STATE{{\bfseries Input:}Downstream task dataset $\mathcal{D}$; \par
        weight matrix of the general-purpose LLM $W^{(0)}$; \par
        the trained parameters of basic medical skills $\Theta$.}
        \STATE Split $\mathcal{D}$ into two parts: $\mathcal{D}^{(adaptation)}$, $\mathcal{D}^{(test)}$.
        \STATE Initialize matrix $A$ and vector $\boldsymbol{b}$ in (\ref{equ:e_matrix}) . \COMMENT{Skill router computation} 
		\IF{normal settings} 
  		    \STATE Optimize loss function in (\ref{equ:normalObjective}) by gradient descent on $\mathcal{D}^{(adaptation)}$;
		\ENDIF
		\IF{few-shot settings}
            \STATE Optimize loss function by CMA-ES on $\mathcal{D}^{(adaptation)}$.
		\ENDIF
        \STATE \textbf{Output:} The skill router $\mathcal{R}$ in (\ref{equ:gate_function}).
        \STATE Fuse the medical skills with the general-purpose LLM by (\ref{equ:fusedModel}). \COMMENT{Fusion}
        \STATE \textbf{Output:} LLM with Medical Domain Knowledge.
        \STATE Inference on $\mathcal{D}^{(test)}$. \COMMENT{Inference}
        \STATE \textbf{Output:} Result $\hat{y}$.
	\end{algorithmic}
\end{algorithm*}

\subsubsection{Fusion and Inference}
After skill adaptation, we get the value of the skill router $\mathcal{R}$. Finally, we fuse each of the skill weight matrix $W_{s_i}$ with the original weight matrix $W^{(0)}$ of the LLM by the skill router. See Figure 4 for the details of fusion process. Assuming that $h$ is an arbitrary hidden layer in the transformer block, the fusion step can be represented as follows:
\begin{equation}\label{equ:fusedModel}
    h = W^{(0)} x + \Delta W x = W^{(0)} x + \sum_{i=1}^{K} \mathcal{R}(\boldsymbol{e}_i) \cdot W_{s_i} x
\end{equation}
where $x$ denotes the input layer. Finally, we use the fused medical LLM to predict the result $\hat{y}$ on the test dataset $\mathcal{D}^{(test)}$. See Algorithm 2 for the process of skill adaptation stage.

%% file: experiment.tex
\subsection{Datasets}
Our proposed knowledge injection framework, SA-MDKIF, consists of two stages. The datasets used in these two stages are described below.

\subsubsection{Skill Training Datasets}
We train 12 skills based on 5 categories of data including: question answering data, text classification data, sequence labeling data, sequence to sequence data, knowledge graph data. The statistics of the corpus used to train the medical skills are shown in Table 1. For skill training, we merge dataset of the same type (e.g., MedQuAD and USMLE). If the same dataset contains more than one skill type, we train them separately (e.g., UMLS is split into EA, ES, and ER skills). 

\begin{table*}[t]
    \centering
    \resizebox{0.9\linewidth}{!}{
        \begin{tabular}{lccc}
            \toprule
            \textbf{Dataset} & \textbf{Data Category}  & \textbf{Skill Type} & \textbf{Corpus Size} \\ \midrule
            MedQuAD~\cite{MedQuAD}   & Question Answering  & MCQA        & 47,457         \\
            USMLE~\cite{jin2021usmle}  & Question Answering   & MCQA        & 61,097        \\
            HealthCareMagic~\cite{li2023chatdoctor}  & Sequence to Sequence    & MC        & 112,165   \\
            UMLS ~\cite{UMLS}  & Knowledge Graph  &  EA + ES + ER   & 15,479      \\
            WikiMed~\cite{wikimed}  & Text Classification   & MLDC   & 393,618    \\
            CliCR~\cite{clicr}  & Sequence Labeling   & MRC        & 10,500      \\
            MIMIC-Cause~\cite{mimicause} & Sequence Labeling  & RE    & 2,714      \\
            MeQSum~\cite{MedQsum}  & Sequence to Sequence  & TS        & 2,333        \\
            EMRQA~\cite{EMRQA}   & Sequence Labeling   & MRC         & 5,789       \\
            PubMedQA~\cite{PubMedQA} & Question Answering   & QA          & 23,149     \\
            MedNLI ~\cite{MedNLI}  & Text Classification   & NLI         & 1,422          \\
            CliNER ~\cite{CliNER}  & Sequence Labeling    & NER         & 1,327         \\ 
            \bottomrule
        \end{tabular}
    }
    \caption{\small Statistics of the dataset used to train basic medical skills. QA: question and answering, MCQA: multiple choice question answering, MC: medical conversation, MLDC: multi-label document classification, MRC: machine reading comprehension, RE: realation extraction, NLI: natural language inference, TS: text summarization, NER: named entity recognition,
    EA: entity attribute, ES: entity synonymy, ER: entity relation.}
    \label{tab:training_skill}
\end{table*}

\subsubsection{Downstream Tasks}
We evaluate our SA-MDKIF over 9 downstream tasks. We divide these tasks into two tracks: \textbf{unseen data} and \textbf{unseen task}, following the work of $\text{MP}^2$~\cite{mp2}. The \textbf{unseen data} track contains 6 datasets that are used in the first stage to train the medical skills, and we keep a small amount of test data $\mathcal{D}^{(test)}$ from the corpus to make sure that the downstream samples are unseen by SA-MDKIF. The \textbf{unseen task} track consists of 3 new downstream medical tasks that are not used during the skill training stage. Table 2 shows the statistics of the downstream medical tasks. \\

\begin{table*}[t]
\centering
\resizebox{0.9\linewidth}{!}{
\begin{tabular}{llcccc}
\toprule
\textbf{Settings}                                                       & \textbf{Downstream Task Data}  & \textbf{Task Type} & \textbf{Test Size} & \textbf{Labels Counts} \\ \midrule
\multirow{6}{*}{\begin{tabular}[c]{@{}l@{}}\textsc{Unseen}\\ \textsc{Data}\end{tabular}} 
& EMRQA~\cite{EMRQA}     & MRC         & 5789           &   N/A              \\
& MedQuAD~\cite{MedQuAD}      & MCQA        & 1273           &   5         \\
& PubMedQA~\cite{PubMedQA}  & QA          & 802            &   N/A        \\
& MIMIC-Cause~\cite{mimicause}    & RE         & 714           &   9         \\
& MedNLI~\cite{MedNLI}    & NLI         & 1422           &   3         \\
& CliNER ~\cite{CliNER}   & NER         & 416            &   N/A         \\ \midrule
\multirow{3}{*}{\begin{tabular}[c]{@{}l@{}}\textsc{Unseen}\\\textsc{Task}\end{tabular}} 
& ICD coding~\cite{mullenbach2018explainable}       & MLDC         & 1584          &   200       \\
& Medication recommendation~\cite{jensen2012mining}    & MLDC         & 582          & 500       \\
& 30-day readmission prediction~\cite{shulan2013predicting}   & DC          & 575          &   2            \\
\bottomrule
\end{tabular}
}
\caption{\small Statistics of downstream tasks. MRC: machine reading comprehension. MCQA: multiple choice question answering, QA: question and answering, RE: relation extraction, NLI: natural language inference, NER: named entity recognition, MLDC: multi-label document classification, DC: document classification.
}
\label{tab:target_task}
\end{table*}

As for an unseen task, we conduct test experiments on the MIMIC-III  dataset~\cite{johnson2016mimic}, a large, open-access database that represents a real-world dataset. The dataset consists of 58,976 admission records for 49,583 patients treated at Beth Israel Deaconess Medical Center between 2001 and 2012. We use its text data for training and testing, and compare the performance of SA-MDKIF with the baseline models on three practical clinical tasks. 
\begin{itemize}
    \item \textbf{ICD coding}~\cite{ICDcoding} is a multi-label classification task based on EHR text for assigning disease labels to patients~\cite{mullenbach2018explainable}.
    \item \textbf{Medication Recommendation}~\cite{jensen2012mining} is a multi-label classification task based on EHR text for automatically recommending medications to patients based on their health conditions
    \item \textbf{30-Day Readmission Prediction}~\cite{shulan2013predicting}  treats patients who are readmitted to the hospital within 30 days of their previous discharge date as positive samples, and it is a binary classification task. This task is of great practical importance in improving the prognosis and quality of patient survival.
\end{itemize}

\subsection{Experimental settings} 
\subsubsection{Data Settings}
We follow previous work to evaluate our model with the \textbf{normal settings}~\cite{normal-settings} and the \textbf{few-shot settings}~\cite{few-shot-settings} of the downstream task data. The normal setting experiment is used to reflect the effect of multiple knowledge sharing and complementarity, while the few-shot setting experiment reflects the model's ability to generalize and transfer learned knowledge to new tasks. 

In the skill adaptation stage, these two settings correspond to different learning strategies. For the normal settings, if the original downstream task has a training set and a test set, we use the original training samples as $\mathcal{D}^{(adaptation)}$, and follow the original test samples as $\mathcal{D}^{(test)}$. Otherwise, we split the complete labeled data into $\mathcal{D}^{(adaptation)}$ and $\mathcal{D}^{(test)}$ in an 8:2 ratio. For the few-shot settings, following the work~\cite{gu2021ppt}, we randomly draw 32 samples from the training set of the downstream task to build an adaptation dataset $\mathcal{D}^{(adaptation)}$ and use the original test samples as $\mathcal{D}^{(test)}$.

\subsubsection{Implementation Details}
We use Llama 2 - 7B~\cite{llama2} as our general-purpose model, which was pre-trained on 2 trillion pieces of data from publicly available sources and released by Meta as an open-source LLM. Llama 2 is an auto-regressive language model that uses an optimized transformer architecture. 

We set the initial rank and target average rank of incremental matrix to 12 and 4, respectively. The orthogonal regularization coefficient is set to 0.5. The number of steps for the initial fine-tuning warm-up and the final fine-tuning are 200 and 1000, respectively. Each skill matrix has a dropout rate of 0.1. The whole process is trained for 6 epochs on 8 A100 GPUs.

\begin{table*}[t!]
\centering
\resizebox{1.0\linewidth}{!}{
\begin{tabular}{llcccccccc}
\toprule
\multicolumn{9}{c}{\textsc{Unseen Data}}\\ \midrule
\multirow{2}{*}{\textbf{Settings}} & \multirow{2}{*}{\textbf{Methods}} & \textbf{Tunable} & \textbf{EMRQA} & \textbf{MeDQuAD} & \textbf{PubMedQA} & \textbf{MedNLI} & \textbf{CliNER} & \textbf{MIMIC-Cause} \\
& & \textbf{Params} & Acc. & Acc. & Acc. & Acc. & F1. & Acc. \\ \midrule
\multirow{8}{*}{\begin{tabular}[c]{@{}c@{}}Normal\\ settings\end{tabular}} 
& Original Llama 2~\cite{llama2}  & 0    & 51.1 & 54.7 & 63.6 & 65.8 & 41.4 & 44.8 \\
& Prompt Tuning~\cite{lester-etal-2021-power} & 3.8M & 60.8 & 65.5 & 80.0 &76.1 & 56.6 & 66.3 \\
& Prefix Tuning~\cite{li-liang-2021-prefix}  & 1.4M & 64.4 & 78.5 & 70.2 & 79.0 & 59.8 & 70.1 \\
& P-tuning ~\cite{Liu2022PTuningPT}  & 6.0M & 58.1 & 76.2 & 85.4 & 75.3 & 52.0 & 71.0 \\
& LoRA ~\cite{hu2021lora}  & 4.2M & 65.3 & 70.7 & 75.4 & 79.1 & 57.9 & 79.3 \\
& IA3 ~\cite{liu2022few}  & 7.0M & 66.7 & 74.0 & 87.3 & 83.9 & 64.2 & 76.9 \\
& Full fine-tuning (Other's)      & 7169M   & \textbf{78.4} & \underline{85.6} & \textbf{94.2} & \textbf{90.9} & \underline{77.9} & \underline{84.0} \\
& $\text{SA-MDKIF}_{-adaptation}$ & 7.2M & 70.9 & 82.0 & 91.7 & 80.1 & 70.4 & 80.1 \\
& $\text{SA-MDKIF}_{normal}$   &7.3M  & \underline{78.0} & \textbf{87.1} & \underline{93.0} & \underline{89.2} & \textbf{79.5} & \textbf{86.9} \\ \midrule
\multirow{3}{*}{\begin{tabular}[c]{@{}c@{}}Few-shot\\ Settings\end{tabular}}  
& Original Llama 2~\cite{llama2}  & 0    & 46.7 & 50.5 & 55.1 & 61.0 & 40.5 & 40.1 \\
& ICL ~\cite{ICL}  & 7.5K & 38.2 & 45.9 & 57.4 & 55.3 & 33.8 & 38.6 \\
& $\text{SA-MDKIF}_{few-shot}$  & 10.9K & \textbf{54.3} & \textbf{61.2} & \textbf{68.8} & \textbf{62.3} & \textbf{50.5} & \textbf{54.6} \\
\bottomrule
\end{tabular}
}
\caption{ \small Comparative experimental results of our SA-MDKIF and other fine-tuning methods in normal and few-shot settings on the Unseen Data. $\text{SA-MDKIF}_{-adaptation}$ represents the ablation version of $\text{SA-MDKIF}_{normal}$ without downstream task adaptation. Few-shot baselines include original Llama 2, In-Context Learing (ICL). Bold font indicates optimal score, underline indicates suboptimal score.}
\label{tab:unseen_data_results}
\end{table*}

\begin{table*}[t!]
\centering
\resizebox{0.8\linewidth}{!}{
\begin{tabular}{llcccc}
\toprule
\multicolumn{6}{c}{\textsc{Unseen Task}}\\ \midrule
\multirow{3}{*}{\textbf{Settings}} & \multirow{3}{*}{\textbf{Methods}} & \textbf{Tunable} & \textbf{ICD} & \textbf{medication} & \textbf{readmission} \\
& & \textbf{Params} & \textbf{coding} & \textbf{recommendation} & \textbf{prediction}\\ 
& & &  AUC. & F1. & AUC. \\ \midrule
\multirow{8}{*}{\begin{tabular}[c]{@{}c@{}}Normal\\ settings\end{tabular}}
& Original Llama 2~\cite{llama2}  & 0    & 29.6 & 37.0 & 50.4 \\
& Prompt Tuning~\cite{lester-etal-2021-power}  & 5.5M & 40.4 & 58.7 & 59.0 \\
& Prefix Tuning ~\cite{li-liang-2021-prefix}   & 1.6M & 41.5 & 55.8 & 63.3 \\
& P-tuning~\cite{Liu2022PTuningPT} & 6.1M & 39.6 & 50.9 & 60.1 \\
& LoRA~\cite{hu2021lora}  & 4.8M & 40.6 & 54.9 & 64.3 \\
& IA3~\cite{liu2022few}   & 7.2M & 51.4 & 60.6 & 70.6 \\
& Full fine-tuning (Other's)     & 7238M   & \textbf{62.3} & \textbf{65.0} & \underline{75.1} \\
& $\text{SA-MDKIF}_{-adaptation}$ & 7.8M & 52.1 & 59.7 & 69.5 \\
& $\text{SA-MDKIF}_{normal}$   & 8.0M & \underline{61.1} & \underline{64.2} & \textbf{76.0} \\
\midrule
\multirow{3}{*}{\begin{tabular}[c]{@{}c@{}}Few-shot\\ settings\end{tabular}}  
& Original Llama 2~\cite{llama2}  & 0    & 25.1 & 30.6 & 43.3 \\
& ICL~\cite{ICL}                 & 24.0K & 27.3 & 33.8 & 49.1 \\
& $\text{SA-MDKIF}_{few-shot}$ & 40.4K & \textbf{39.8} & \textbf{40.5} & \textbf{52.4}\\
\bottomrule
\end{tabular}
}
\caption{\small Comparative experimental results of our SA-MDKIF and other fine-tuning methods in normal and few-shot settings on the Unseen Task. $\text{SA-MDKIF}_{-adaptation}$ represents the ablation version of $\text{SA-MDKIF}_{normal}$ without downstream task adaptation. Few-shot baselines include original Llama 2, In-Context Learing (ICL). Bold font indicates optimal score, underline indicates suboptimal score.}
\label{tab:unseen_task_results}
\end{table*}

\subsection{Results and Analysis} 
The main results of the comparative experiments on 9 medical tasks are presented in Table 3 and Table 4. Overall, our SA-MDKIF method outperforms the other PEFT methods and demonstrates its superiority. The results we report are the average performance over 5 runs with different random seeds. 

\subsubsection{Normal settings} 
(1) \textbf{Comparison with the original Llama 2.} 
The results of all PEFT methods on the 6 datasets show significant performance improvements over the original Llama 2, demonstrating the importance of medical knowledge injection for general-purpose LLM. \\
(2) \textbf{Comparison with the other PEFT methods.} 
Compared to the 5 main PEFT methods, our SA-MDKIF method achieves the best results on most of the datasets, further demonstrating the advantage of skill adaptation for downstream tasks.\\
(3) \textbf{Comparison with full fine-tuning.}
For all 9 test datasets in Unseen Data and Uneen Task, SA-MDKIF outperforms the full fine-tuning method for 4 of them, and is comparable to it for the other 5 datasets, but with only about 1\% of the number of parameters, demonstrating the efficiency and performance of our method. \\
(4) \textbf{Comparison with only one stage.} We constructed the $\text{SA-MDKIF}_{-adaptation}$ version without the downstream adaptation, where all medical skills are equally weighted. $\text{SA-MDKIF}_{-adaptation}$ performs worse than SA-MDKIF on either \textit{unseen data} or \textit{unseen task}, demonstrating the rationality and effectiveness of the two-stage framework we have designed. It is worth noting that $\text{SA-MDKIF}_{-adaptation}$ outperforms LoRA in most datasets, demonstrating the advantage of AdaLoRA during the skill training stage.

\subsubsection{Few-shot settings} 
\textbf{Overall Performance.} We compared SA-MDKIF with the original Llama 2 and In-Context Learning (ICL) baselines in the few-shot settings. The experimental data in Table 3 and Table 4 show that $\text{SA-MDKIF}_{few-shot}$ has higher accuracy compared to the original Llama 2, while outperforming the overall performance of In-Context Learning (ICL) in few-shot learning. Importantly, our method is close to zero-shot in terms of the number of tokens used, but significantly less than ICL. In the era of LLM, where the input length is proportional to the inference cost, the ability to achieve near-peak performance using the smallest possible number of tokens becomes increasingly important.\\

\begin{figure}[htbp]
    \centering
    \includegraphics[width=0.5\textwidth]{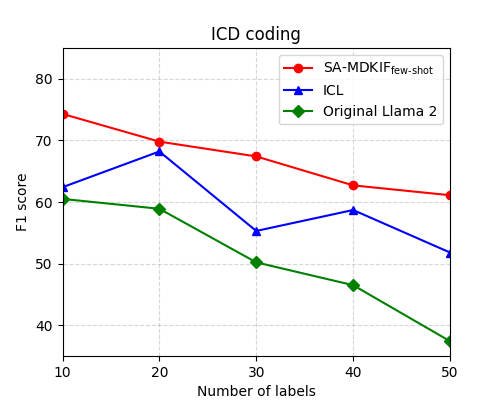}
    \caption{F1 scores of the few shot learning methods comparison on different number of labels in a multi-label classification task (ICD coding).}
    \label{fig5}
\end{figure} 

\textbf{On Multi-Label Classification Tasks.} Our proposed method, SA-MDKIF, uses a \textit{unified instruction format} in both skill training stage and skill adaptation stage. Thus, it can handle tasks with different numbers of labels. To test the performance of the model on different numbers of labels, we take the ICD coding task as an example. We compare the F1 values of SA-MDKIF and the other two baseline models for the 10/20/30/40/50 labels with the highest frequency of occurrence. As can be seen in Figure 5, there is a sharp drop in the F1 score of the ICL when the number of labels is greater than 20. In contrast, the performance of SA-MDKIF declines more slowly and steadily as the number of labels increases. This demonstrates the superiority of the \textit{unified instruction format}.

\subsubsection{Comparison with downstream baseline models}
To ensure fairness, we use the data from MIMIC-III to train a new skill and plug it into LLM to get a new version of $\text{SA-MDKIF}_\text{new}$. Then, we compare the performance of $\text{SA-MDKIF}_\text{new}$ with the baseline models on the test data using the three clinical tasks mentioned above: ICD coding, medication recommendation, and readmission prediction.

\begin{table*}[t!]
\centering
\resizebox{0.8\linewidth}{!}{
\begin{tabular}{llccc}
\toprule
\textbf{Task} & \textbf{Model} & \textbf{Accuracy\%} & \textbf{AUC\%}  & \textbf{F1\%} \\
\midrule
\multirow{5}{*}{\shortstack[l]{ICD \\ coding \\~\cite{mullenbach2018explainable}}} 
& CNN~\cite{shickel2017deep}  & 32.53 & 83.59 & 49.08 \\
& ClinicalBERT~\cite{huang2019clinicalbert} & 33.01 & 84.11 & 49.65\\
& ClinicalBERT+G-BERT~\cite{ijcai2019p825}  & 33.46 & 85.76 & 50.14 \\
& MedM-PLM~\cite{MedM-PLM} & 34.89 & 86.78 & 52.03 \\
& $\text{SA-MDKIF}_{new}$  & \textbf{40.27} & \textbf{90.37} & \textbf{57.12} \\
\midrule
\multirow{5}{*}{\shortstack[l]{Medication \\ Recommendation \\~\cite{jensen2012mining}}} 
& LR & 88.97 & 77.34 & 61.38 \\
& RNN~\cite{connor1994recurrent} & 90.51 & 91.85 & 58.50 \\
& Med-BERT~\cite{rasmy2021med} & 91.01 & 93.04 & 61.47 \\
& MedM-PLM~\cite{MedM-PLM} & 92.03 & 95.38 & 67.21 \\
& $\text{SA-MDKIF}_{new}$ & \textbf{94.65} & \textbf{96.57} & \textbf{72.41} \\
\midrule
\multirow{4}{*}{\shortstack[l]{Readmission \\ Prediction \\~\cite{shulan2013predicting}}} 
& CNN~\cite{shickel2017deep} & 62.55 & 66.68 & 57.83 \\
& ClinicalBERT~\cite{huang2019clinicalbert} & 64.07 & 69.33 & 63.50 \\
& MedM-PLM~\cite{MedM-PLM} & 68.74 & 74.03 & 68.61 \\
& $\text{SA-MDKIF}_{new}$  & \textbf{71.12} & \textbf{79.36} & \textbf{75.32}\\
\bottomrule
\end{tabular}
}
\caption{\small Performance comparison of SA-MDKIF and baseline models on three clinical tasks. All results are averaged over five runs, with optimal results in bold.}
\label{tab:comp_with_sota}
\end{table*}

From the results demonstrated in Table 5, it can be seen that the AUC and F1 scores of $\text{SA-MDKIF}_\text{new}$ on all three clinical tasks exceeded the corresponding scores of the Bert-based baseline models. In addition, the performance metrics of our $\text{SA-MDKIF}_\text{new}$ outperform the results of the unseen task shown in Table 4 due to the training of the new skill based on MIMIC-III training data. The above results confirm the superior performance and architectural scalability of SA-MDKIF.

\subsubsection{Effect of skill number on performance}
We selected three tasks, EMRQA, medication recommendation, and readmission prediction, to observe the effect of different numbers of skills on the performance of our SA-MDKIF by adjusting the number of skills involved in the training.\\

\begin{figure}[htbp]
    \centering
    \includegraphics[width=0.5\textwidth]{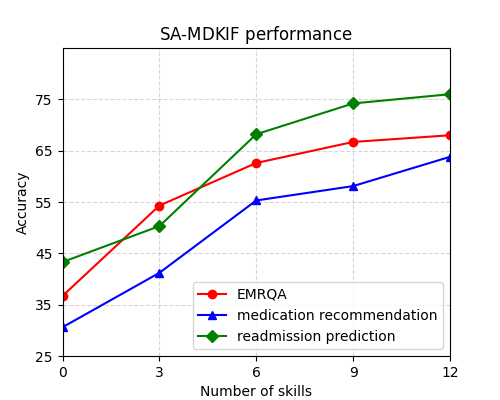}
    \caption{Effect of skills number on SA-MDKIF performance.}
    \label{fig6}
\end{figure}

Figure 6 shows the results of the experiment, where we can see that our SA-MDKIF can continuously improve its performance as the number of skills increases. Moreover, with only 3 skills, SA-MDKIF performs much better than the original Llama 2, which well validates the advantage of our proposed knowledge injection framework.

%% file: reference.tex
\bibliography{reference}

%% file: appendix.tex
\section{Details of AdaLoRA}
We use $j$ to index the incremental matrix, i.e.,  $ \Delta_{j} = U_j \Lambda_j V_j $ where $j \in \Gamma=\{q,k,v,f_1,f_2,o\} $. We further donate the parameter sets $ \mathcal{U} = \{U_j\}_{j \in \Gamma}, \mathcal{E} = \{\Lambda_j\}_{j \in \Gamma} , \mathcal{V} = \{V_j\}_{j \in \Gamma} $. Then the final loss function of a specific skill $s_i$ can be described as follows:
\begin{equation}\label{equ:skill_loss}
   \mathcal{L}(\mathcal{U},\mathcal{E},\mathcal{V}) = \mathcal{C}(\mathcal{U},\mathcal{E},\mathcal{V}) + \gamma \sum_{j \in \Gamma} R(U_j,V_j) 
\end{equation}
\begin{equation}\label{equ:regularizer}
    R(U,V) = ||U^TU-I||^2_F+||V^TV-I||^2_F
\end{equation} 
where $\mathcal{C}(\mathcal{U},\mathcal{E},\mathcal{V})$ denotes the loss function on the training data, $R(U,V)$ 
is the regularizer to enforce the orthogonality of $U$ and $V$, $\gamma$ is the hyper-parameter. \\

In order to control the budget of the fine-tunable parameters, the trainable parameters are dynamically assigned during the training process. At the $t$-th step, where $t$ is between the initial fine-tuning warm-up step $t_0$ and the final fine-tuning step $t_1$, we take a stochastic gradient step to update $U_j^{(t)}$, $\Lambda_j^{(t)}$, $Q_j^{(t)}$ for $j \in \Gamma$. Specifically, for $\Lambda_j^{(t)}$:
\begin{equation}\label{equ:gradient_update}
\hat{\Lambda}_j^{(t)}=\Lambda_j^{(t)}-\eta\nabla_{\Lambda_j} \mathcal{L}(\mathcal{U}^{(t)},\mathcal{E}^{(t)},\mathcal{V}^{(t)})
\end{equation}
where $\eta$ denotes the learning rate.
We further donate $\mathcal{G}_p = \{U_{*p}, \Lambda_{p}, V_{p*}\}$ as the triplet containing the $p$-th sigular value and vectors. The gradient is then trimmed according to the importance of each triplet to obtain $\hat{\Lambda}_{j}^{(t+1)}$, retaining only the singular values whose importance satisfies the requirement. Given the importance score $S_j^{(t)}$, the sigular values are pruned as follows:
\begin{equation}\label{equ:gradient_trimming}
\hat{\Lambda}_{j}^{(t+1)} = \mathcal{F} (\hat{\Lambda}_j^{(t)},S_j^{(t)})
\end{equation}

\begin{equation}\label{equ:budget_schdular}
    \mathcal{F} (\hat{\Lambda}_j^{(t)},S_j^{(t)})_{pp} = \begin{cases}
    \hat{\Lambda}_{j,pp}^{(t)} & S_{j,p}^{(t)} \text{ is in the top-}b^{(t)} \text{ of } S^{(t)} , \\
    0 & \text{ otherwise},
    \end{cases}
\end{equation}
where $S^{(t)} = \{S_{j,p}^{(t)}\}_{1 \leq j \leq m, 1 \leq p \leq r}$ contains the importance scores of all triplets, $b^{(t)}$ is the budget of remaining singular values at the $t$-th step. The importance of a particular triplet is computed as follows:
\begin{equation}\label{equ:score_function}
S_{j,p} = s(\lambda_{j,p}) + \frac{1}{d_1} \sum_{q=1}^{d_1} s(U_{j,qp}) + \frac{1}{d_2} \sum_{q=1}^{d_2} s(V_{j,pq})
\end{equation}
where $S_{j,p}$ denotes the importance score of the $p$-th triple of the $j$-th weight matrix, $\lambda_{j,p}$ denotes the $p$-th element of the $\Lambda$ matrix of the $j$-th weight matrix, $U_{j,qp}$ denotes the $(q,p)$ element of the $U$ matrix of the $j$-th weight matrix, and $V_{j,pq}$ denotes the $(p,q)$ element of the $V$ matrix of the $j$-th weight matrix. $s(.) $ denotes the importance of a parameter, which is defined as the magnitude of any trainable parameter $w_{pq}$ and its gradient $\nabla_{w_{pq}} \mathcal{L}$ : $s(w_{pq}) = | w_{pq} \nabla_{w_{pq}} \mathcal{L}|$. This formula approximates the change in the loss function when the parameter becomes zero, meaning that if a parameter is cropped and the loss function changes significantly, we should keep it.
For more detailed explanation of the formulas, please refer to the AdaLoRA~\cite{AdaLoRA} paper. \\